# Retraining as Approximate Bayesian Inference

HARRISON KATZ

**PREVIEW** *Model retraining is usually treated as an ongoing maintenance task. But as Harrison Katz now argues, retraining can be better understood as approximate Bayesian inference under computational constraints. The gap between a continuously updated belief state and your frozen deployed model is "learning debt," and the retraining decision is a cost-minimization problem with a threshold that falls out of your loss function. In this article Katz provides a decision-theoretic framework for retraining policies. The result is evidence-based triggers that replace calendar schedules and make governance auditable. For readers less familiar with the Bayesian and decision-theoretic language, key terms are defined in a glossary at the end of the article.*

### THE BAYESIAN FRAMEWORK VS. RETRAINING

In Bayesian inference, learning is continuous. A model maintains a distribution over parameters, not a point estimate. As data arrive, the distribution updates continuously: prior beliefs combine with evidence to form a posterior. If conditions are stable, updates are small. If conditions change, the posterior shifts as evidence accumulates. There is no moment at which the model becomes stale – the model is never "old" or "new." It is simply the current posterior, conditioned on all observed data.

This is the key insight that makes "When should we retrain?" a strange question within the Bayesian framework. Here, the concept of *retraining* doesn't occur. If the data-generating process is stable, updates are incremental. If the process changes, the posterior adapts as evidence accumulates. There is no calendar event at which the system becomes wise again.

However, the Bayesian ideal is computationally expensive. Exact updating is often intractable for large models. Many production systems ship point estimates rather than full posteriors. Organizations face deployment pipelines, validation requirements, and governance constraints that make continuous updating impractical.

So in practice, we are forced to batch. We accumulate data, retrain periodically, validate, and deploy. But this discrete cadence is a property of compute and process, not of learning itself (Gama et al., 2014).

Once we batch, we create a decision problem: When is it worth paying the cost of a new training run? The traditional framing treats this as maintenance: refresh the model every week, every month, or whenever a drift alarm fires. An example of such an alarm, using control charts to detect drift, appeared in *Foresight* Issue 56 (Katz, 2020).

The control chart approach triggers retraining when residual patterns are statistically unusual under the assumption of a stable process. The Bayesian approach triggers when the expected cost of a stale model exceeds the cost of intervention. The first asks "Is this surprising?" The second asks "Is acting worth it?" Both approaches replace rigid schedules with evidence-based triggers, but the basis for setting the trigger differs.

### VIEWING TRAINING AS DEBT REDUCTION

In this article I propose a different framing. Consider what we are trying to approximate. With infinite computational resources and zero operational friction, we would maintain a continuously updat-

> **Key Points**
>
> - Retraining is not maintenance. It is approximate Bayesian inference under computational and operational constraints.
> - The gap between your deployed model and a hypothetical continuously updated model is "learning debt." Monitoring metrics should estimate this debt, not just track error.
> - The retraining decision has asymmetric costs: churn (retraining when stable) and bias (not retraining when shifted). The optimal threshold follows from those costs.
> - Thresholds should be derived from your loss function, not picked because they seem reasonable or match historical cadence.

ed belief state. Call this the *continuously updated posterior*.

In practice, the deployed model reflects the posterior from the last training event. Between training events, data arrive but the belief state stays frozen. The gap between the belief we would have under continuous updating and the belief we actually deploy is *accumulated learning debt*, analogous to technical debt in machine learning systems (Sculley et al., 2015).

In information-theoretic terms, learning debt is a divergence between two belief states. KL divergence (a measure of how one probability distribution differs from a reference probability distribution) is a natural choice (Kullback and Leibler, 1951). We do not need to compute this divergence exactly. What matters is that we can build proxies that track it and then use those proxies to decide when an intervention is justified. We refer to this approach as the *learning debt framework*.

### THE DECISION-THEORETIC LAYER

The problem is not only that the model may be stale. We are also uncertain about *why*. Is the world changing, or did we observe noise? Is the shift temporary or permanent?

A useful formalization is to maintain a belief about whether the data-generating process has changed. Call this P(shift). It can be explicit (a Bayesian model over regimes) or implicit (a calibrated drift score).

This shift probability need not be constant. Regime changes are more likely around known interventions (product launches, pricing changes, policy shifts) and macro shocks. Bayesian online changepoint detection formalizes this with a hazard rate, a prior on regime change at each step (Adams and MacKay, 2007). A practical approximation is to raise the shift prior during turbulent windows and demand stronger evidence during quiet periods.

Retraining then becomes a decision problem with asymmetric costs:

- Retrain when stable: pay churn cost (compute, deployment risk, potential regressions).
- Do not retrain when shifted: pay bias cost (accumulating forecast error until you catch up).
- Retrain when shifted: pay retrain cost, avoid most bias cost.

Standard decision-theoretic reasoning (Berger, 1985) yields a simple inequality:

*Retrain when P(shift) > (churn cost) / (bias cost)*

This is where practitioner guidance often fails. Teams pick thresholds because they seem reasonable or match historical cadence. The decision-theoretic framing forces a useful question: What are the actual costs of acting too often versus too late? Once we write those costs down, the threshold is no longer arbitrary. It's a design parameter grounded in the loss function.

### PROXIES FOR POSTERIOR DIVERGENCE

We cannot compute the continuously updated posterior in most production systems. If we could, we would not be batching. But we can build approximations that aim at the right target.

The key change in mindset is to treat monitoring metrics as measurements of belief staleness, not just performance degradation.

### Predictive evidence on fresh data

The most direct signal is how surprised the deployed model is by recent outcomes. Compute proper scoring rules (log loss, CRPS) on a rolling window of fresh data (Gneiting and Raftery, 2007; Hersbach, 2000). Systematic surprise indicates a stale belief state. In Bayesian terms, repeated surprise is evidence that the frozen posterior no longer aligns with observed data.

### Calibration and distributional mismatches

Many models maintain acceptable average error while becoming miscalibrated in ways that matter downstream. Track calibration curves, prediction interval coverage, and group-level residual structure. Posterior predictive checks provide a useful mindset: compare what the model implies to what the data show, and quantify the gap (Gelman, Meng, and Stern, 1996). These checks are interpretable. They indicate not just that the model is worse, but how. For example, calibration curves reveal whether the model overestimates or underestimates in specific regimes. Group-level residuals show which customer segments or product categories are drifting. These diagnostics guide not just the decision to retrain, but what to fix.

### Shadow fits and parameter stability

Fit a lightweight model on a recent window and compare it to the deployed model. This shadow model need not match the full production architecture. If the shadow wants very different parameters, that is evidence of drift (Gama et al., 2014; Bifet and Gavalda, 2007). The disagreement between deployed and shadow models estimates learning debt.

This approach separates two failure modes. If both models perform poorly, the issue may be features, labels, or pipeline rather than drift. If the shadow improves materially over the deployed model, the world has moved and we are behind.

### Domain-specific distributional signals

Some forecasting systems fail not because demand levels changed, but because timing or composition changed. In travel, a pickup forecast may break because lead-time distributions compressed. In retail, product mix may shift while aggregate demand stays flat. In ad bidding, auction competition distributions may change faster than conversion rates.

In each case, tracking distributional divergence (L1, KL, Wasserstein) on the relevant domain object serves as a sensitive early indicator of drift. In the framing of this article, such divergence is a domain-specific proxy for learning debt.

## IMPLEMENTATION

Real models are not one-parameter rates. We usually cannot compute exact beliefs. But we can implement the logic as a policy layer around the training pipeline.

### Architecture

- **Deployed model:** production model
- **Fresh-data evaluator:** computes proper scoring rules on a rolling window.
- **Shadow learner:** lightweight model retrained frequently, or a calibrator/last-layer update.
- **Evidence aggregator:** converts performance differences into an evidence score for stable versus shifted.
- **Policy threshold:** compares evidence-adjusted shift belief to the cost ratio, triggers retraining when justified.

### Policy Template

1. Define what a shift means in the domain: outcomes, inputs, or timing structure.
2. Choose two or three evidence signals that approximate belief staleness.
3. Write down churn cost and bias cost in common units. Run sensitivity analysis.
4. Set the threshold implied by those costs. Implement as code, not lore.
5. Backtest on known disruptions and quiet periods to validate.

Notice what is absent: a universal drift threshold. The threshold is a policy parameter derived from operational reality. If deployments are risky, set the bar higher. If prediction error is expensive and fast moving, set it lower.

This makes governance easier. When asked why we retrained, we answer in auditable terms: the evidence for a real shift crossed the decision threshold implied by our costs.

A full implementation guide is beyond the scope of this article. The architecture above provides a starting point. Details will vary by tech stack, governance requirements, and organizational context.

### Choosing Costs

Cost models need not be perfect to be useful. Rough numbers are better than implicit numbers in people's heads.

One approach is to define costs in the same units as the business objective.

- **Churn cost:** engineering hours, compute, deployment risk, translated into dollars or utility.
- **Bias cost:** expected downstream loss per day of stale predictions times expected duration until intervention.
- **Retrain cost:** marginal cost of training, evaluation, and deployment.

For instance, churn cost might include four engineer-hours per retrain at $150/hour, plus a deployment risk premium based on historical regression rates. Bias cost might be $500 per day of degraded forecasts, derived from downstream inventory holding costs or revenue impact from mispriced promotions. Exact numbers matter less than order of magnitude. If churn cost is around $1,000 and bias cost is $500/day, the threshold implies retraining is worthwhile if you expect the model to remain stale for more than two days.

Then run sensitivity analysis. If the policy changes only under extreme assumptions, the decision is robust. If it flips under small changes, that identifies where better measurement pays off. In practice, sensitivity analysis is a few hours of spreadsheet work: vary the cost assumptions by 2x in each direction and see whether the policy flips. This is not a major undertaking.

### Stylized Examples

*Travel demand*

Consider a travel demand forecasting system. Each morning we observe on-the-books reservations for future check-in dates and add an expected pickup curve to predict final totals. At the last training event, the model implied that by 30 days before arrival, 60% of eventual demand is already booked.

Now booking windows shift. Travelers book later. At 30 days out, only 40% of eventual demand is booked. The feature pipeline is correct. The data-generating process for timing has changed.

Waiting for final outcomes reveals this too late. But tracking distributional divergence between the current lead-time histogram and a baseline from the training era detects the change immediately. Distribution-level monitoring catches changes that are missed by averages and medians.

In the language of this article, sustained lead-time divergence is evidence that the deployed belief state is stale. A shadow timing model fit on recent weeks serves as the continuously updated alternative we approximate. The action rule is unchanged: retrain when expected cost of stale forecasts, weighted by P(shift), exceeds churn cost.

*Retail promotion response*

Consider a retail demand forecasting system for a consumer product. The model was trained on two years of history including seasonal patterns and promotional lifts. At the last training event, the model learned that a 20% price promotion typically generates a 2.5x demand multiplier.

Now the competitive environment shifts. A new entrant runs frequent promotions, eroding the lift from your own discounts. The same 20% promotion now generates only a 1.6x multiplier. Aggregate demand may look stable, but the promotional response has changed.

Tracking point forecasts alone may miss this. But monitoring the distribution of forecast errors conditional on promotion status reveals the gap immediately. The model systematically overpredicts during promotions and underpredicts during regular periods. In the language of this article, the promotional response model has accumulated learning debt. A shadow model fit on recent promotions estimates what the updated belief would be. The action rule is unchanged: retrain when expected cost of biased promotional forecasts exceeds churn cost.

*A note on temporary vs. permanent shifts*
The learning debt framework does not require knowing in advance whether a shift is temporary or permanent. It responds to evidence as it accumulates. If the shift reverses quickly, the divergence signal fades and no intervention is triggered. If it persists, evidence accumulates until the cost threshold is crossed. The model that was "correct" before the shift is not assumed to be perfect – only that it represented the best available belief at the time. What matters is whether the current deployed belief has drifted far enough from what continuous updating would yield to justify the cost of intervention.

## LIMITATIONS

The learning debt framework assumes retraining is a meaningful discrete event with nontrivial cost. It is less useful in two boundary cases.

First, when updates are nearly continuous. High-frequency systems (ad bidding, recommendations, trading) sometimes update on every batch. The question "When to retrain?" dissolves into "How much to learn from each observation," better handled by learning rate schedules and online learning theory.

Second, when the cost ratio is unknowable. Some organizations cannot estimate bias cost because downstream effects are diffuse or contested. The framework still helps by making disagreement explicit, but it will not resolve it. We can identify what must be true for different policies to be justified and let stakeholders argue assumptions rather than thresholds.

This framework also assumes we can build proxies cheaper than full retraining. If the only way to know whether the model is stale is to retrain and compare, the monitoring layer adds overhead without decision value. In practice this is rare. Most systems have cheap signals (scoring rule degradation, shadow disagreement, distributional shift) that inform before full retraining cost is incurred. The monitoring layer is lightweight: scoring rules and distributional comparisons scale linearly with the number of series. For systems with thousands or millions of time series, the marginal cost of monitoring is negligible. The expense is in the retraining itself, which this framework aims to trigger less often, not more.

This framework requires enough history to establish stable baselines for the monitoring signals and to estimate costs with reasonable confidence. In practice, this is typically whatever was sufficient to train the original model. Systems with very short histories may need to rely on domain priors for cost estimates until enough data accumulates.

## CONCLUSION

The learning debt framework offers several advantages:

- It replaces calendar schedules with evidence and costs.
- It turns drift detection from a grab bag of metrics into a coherent attempt to estimate belief staleness.
- It handles predictable turbulence by design. The shift prior can be higher around launches, policy changes, and calendar events (Adams and MacKay, 2007).
- It makes thresholds defensible, auditable, and tunable to risk tolerance.
- It clarifies what retraining buys: not novelty, but reduced learning debt (Wald, 1950; Berger, 1985).

Production ML is not only a modeling problem. It is a sequential decision problem under uncertainty. The Bayesian ideal supplies the conceptual target. Decision theory supplies the action rule. Everything in between is engineering: finding the cheapest, safest approximations that move toward the ideal.

*Glossary of Bayesian and Decision-Theoretic Terms*

**Belief state:** The current probability distribution representing what the model "knows." In production systems, this is typically frozen between training events.

**Calibration:** A probabilistic forecast is well calibrated if events predicted with X% probability occur X% of the time.

**CRPS (Continuous Ranked Probability Score):** A proper scoring rule for probabilistic forecasts that generalizes mean absolute error to full predictive distributions.

**Data-generating process (DGP):** The underlying mechanism that produces observed data. A regime shift means this mechanism has changed.

**Hazard rate:** In changepoint detection, the prior probability that a regime shift occurs at any given time step.

**KL divergence:** A measure of how different two probability distributions are. Used here to quantify learning debt.

**Learning debt:** Accumulated divergence between a continuously updated belief and the deployed (frozen) belief. Analogous to technical debt in software systems.

**Posterior:** Updated belief about model parameters after observing data. Combines the prior with observed evidence.

**Posterior predictive:** Distribution of future outcomes implied by the model and its uncertainty.

**Prior:** Belief about model parameters before observing data.

**Proper scoring rule:** Metric that rewards accurate probabilistic predictions. Examples include log loss and CRPS. Unlike point-forecast metrics, proper scoring rules incentivize honest uncertainty estimates.

**Regime shift:** Change in the data-generating process that makes older data less informative.

**Shadow model:** A lightweight model trained on recent data, used to estimate what parameters a continuously updated system would have learned.

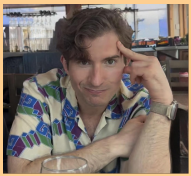

**Harrison Katz** holds a PhD in statistics from UCLA. He is Head of Finance Data Science & Strategy at Airbnb, leading the forecasting team that supports earnings, treasury operations, and strategic planning and advising on enterprise data strategy. His research focuses on Bayesian methods for compositional and hierarchical time series, including the B-DARMA and B-DARCH frameworks for lead-time and volatility forecasting. Prior to Airbnb, he held research positions at the Federal Reserve Board of Governors.

harrison.katz@airbnb.com